\documentclass[article]{acmart}

\AtBeginDocument{%
	}

\setcopyright{acmlicensed}
\copyrightyear{2024}
\acmYear{2024}
\acmDOI{XXXXXXX.XXXXXXX}

\acmJournal{JACM}
\acmVolume{37}
\acmNumber{4}
\acmArticle{111}
\acmMonth{8}

\usepackage{array}
\usepackage{booktabs} 
\usepackage{diagbox}
\usepackage{multirow}

\begin{document}
	
\title{Neuro-Symbolic AI: Explainability, Challenges, and Future Trends}
	
\author{XIN ZHANG}
\orcid{0000-0003-4960-174X}
\affiliation{%
	\institution{Department of Computer Science, Texas Tech University}
	\streetaddress{2500 Broadway}
	\city{Lubbock}
	\state{Texas}
	\postcode{79409}
	\country{USA}}
\email{e-mail: zha19053@ttu.edu}

\author{VICTOR S.SHENG}
\orcid{0000-0003-4960-174X}
\affiliation{%
	\institution{Department of Computer Science, Texas Tech University}
	\streetaddress{2500 Broadway}
	\city{Lubbock}
	\state{Texas}
	\postcode{79409}
	\country{USA}}
\email{e-mail: victor.sheng@ttu.edu}

	\renewcommand{\shortauthors}{Trovato et al.}
	
\begin{abstract}
Explainability is an essential reason limiting the application of neural networks in many vital fields. Although neuro-symbolic AI hopes to enhance the overall explainability by leveraging the transparency of symbolic learning, the results are less evident than imagined. This article proposes a classification for explainability by considering both model design and behavior of 191 studies from 2013, focusing on neuro-symbolic AI, hoping to inspire scholars who want to understand the explainability of neuro-symbolic AI. Precisely, we classify them into five categories by considering whether the form of bridging the representation differences is readable as their design factor, if there are representation differences between neural networks and symbolic logic learning, and whether a model decision or prediction process is understandable as their behavior factor: implicit intermediate representations and implicit prediction, partially explicit intermediate representations and partially explicit prediction, explicit intermediate representations or explicit prediction, explicit intermediate representation and explicit prediction, unified representation and explicit prediction. We also analyzed the research trends and three significant challenges: unified representations, explainability and transparency, and sufficient cooperation from neural networks and symbolic learning. Finally, we put forward suggestions for future research in three aspects: unified representations, enhancing model explainability, ethical considerations, and social impact.

\end{abstract}

	\ccsdesc[500]{Do Not Use This Code~Generate the Correct Terms for Your Paper}
	\ccsdesc[300]{Do Not Use This Code~Generate the Correct Terms for Your Paper}
	\ccsdesc{Do Not Use This Code~Generate the Correct Terms for Your Paper}
	\ccsdesc[100]{Do Not Use This Code~Generate the Correct Terms for Your Paper}
	
	\keywords{Do, Not, Us, This, Code, Put, the, Correct, Terms, for,
		Your, Paper}
	
	\received{20 September 2024}
	\received[revised]{12 September 2024}
	\received[accepted]{5 September 2024}
	
	\maketitle
	
\section{Introduction}
The difficulties of explaining neural networks come from its feature extraction and reasoning process based on features. It could be more intuitive to associate float numbers with our macroscopically perceived object features. Although many attempts have been made to explore this aspect, it still takes much effort to solve this problem fundamentally. Explainability is less evident than other metrics for measuring model performance, such as accuracy, recall or precision.

\citet{ref1facchin2023neural} proposed that researchers have yet to directly observe or identify a clear neural structure that fully meets or represents a specific NSR (neural structure representation) standard in neuroscience. In other words, although significant progress has been made in understanding how the brain processes and stores information, we still need to be able to specify an exact neural structure or network of neurons that precisely corresponds to a specific encoding of information. In addition, \citet{ref2van2023cleaning} discovered that the significant correlation between computational models and the human brain does not prove they implement the same process. 

\citet{ref3yang2022psychological} pointed out that the current field of XAI (explainable artificial intelligence) needs a precise computational theoretical framework to explain how humans understand the explanation of AI, which limits the development of theoretical construction of XAI systems. They argue that humans expect AI to make decisions independently and understand AI's explanations by comparing them to their knowledge. This comparison process can be formalized using the universal generalization laws discovered by Shepard in similar spaces. \citet{ref197shepard1987toward} proposed a theoretical framework to predict how humans conclude AI explanations, achieving a more accurate and effective explanation method consistent with human cognitive processes. \citet{ref4fedyk2023leverage} observed that many efficient machine-learning algorithms belong to the "Rashomon set." Although these models have similar performance indicators, they may differ significantly in structure or interpretation. This diversity suggests that even though some current models, as analogs of human cognitive processes, may provide a basis for generating hypotheses about human cognition, it is difficult to make accurate predictions about them because of current explainability limitations.

\citet{ref5broniatowski2021psychological} illustrated the different but complementary roles that explainability and understandability play in making machine learning systems more transparent and trustworthy to users. The two are not mutually exclusive but provide solutions for different aspects of user interaction with machine learning systems. Interpretability focuses on describing in detail the internal mechanisms or processes by which a model makes specific decisions to allow the model's designers or professionals to understand and verify the model's behavior. Comprehensibility emphasizes the meaning of model's output to users, that is, how to transform the output into information or a decision-making basis that is of practical significance to users. \citet{ref6verhagen2022exploring} proposed from the cognitive psychology perspective that special attention should be paid to the four components of perception, semantics, intention, user, and context when designing explanations. This analysis means that an additional process may need to be introduced between the original explanation generated by the AI system and the explanation ultimately provided to the user, as the former may need to be more technical and complex to understand for non-expert users. This step transforms the original explanation into a more humane and easy-to-understand form. This research inspires us to balance the technical transparency of the model with the explainability of user expectations.

The above studies show that we should be cautious about the explainability of neuro-symbolic AI, even though the logic of symbolic learning is often transparent and readable. This survey analyzes 191 neuro-symbolic AI studies from the past decade to provide research references on explainability for scholars who wish to conduct related research. Chapter two starts with an analysis of the models' explainability and then divides these studies into five categories from the perspective of design and behavior. Chapter three proposes a more detailed classification description and statistics with 14 representative cases for discussion. Then, Chapter four briefly analyzed the general trends in neuro-symbolic AI based on collected articles. Next, from Chapters Five to Six, we analyze the main research challenges and promising future research directions. 
\section{Explainability Analysis}	
\citet{ref7calegari2020integration} divided models' transparency into explainability by design and post-hoc. The former designs the models to be easily understood by humans or embeds logical rules and constraints in the model construction process to make the neuro-symbolic AI system transparent and explainable by design, essentially avoiding the impact of black box models from the beginning. The latter refers to analyzing the model's behavior to explain its decision-making process after the system has been designed, developed, and deployed, such as by generating explanatory text, visualization technology, and model simplification.

Combining these two ideas, we proposed an explainability classification method that considers both the design and behavior of the models. As Table 1 shows, we utilize this method to classify 191 research into five categories: low, medium-low, medium, medium-high, and high.

First, we consider the difference between features extracted by neural networks and the form of information processed by symbolic logic. This perspective focuses on the compatibility and conversion mechanism between these two diagrams. In addition to the explainability of neural network itself, the format of this conversion also affects the explainability of integrated models, contributing to the first standard for categorization, which measures the readability of intermediate representation that bridges the neural representation and the logical symbolic representation. Second, we consider the explainability of decision-making or prediction logic in neuro-symbolic AI models. Specifically, even considering the inevitable impact of the black-box processing of neural networks, we still can understand the essence of knowledge-processing methods to various degrees, which will help us to explain the decision-making or prediction to a certain degree. Based on the above conditions, our specific classification method is shown in Figure 1.

\begin{table}[ht]
	\centering 
	\caption{Neuro-Symbolic AI's classification by explainability} 
	\label{tab:neuro_symbolic}
	\begin{tabular}{
			>{\centering\arraybackslash}m{2.3cm} 
			>{\centering\arraybackslash}m{2.15cm} 
			>{\centering\arraybackslash}m{2.15cm}
			>{\centering\arraybackslash}m{2.15cm} 
			>{\centering\arraybackslash}m{2.15cm} 
			>{\centering\arraybackslash}m{2.15cm} 
		}
		\toprule 
		Explainability & Implicit Intermediate Representations & Partially Explicit Intermediate Representations & Explicit Intermediate Representations & Explicit Intermediate Representations & Unified Representation \\
		\midrule 
		Implicit Decision-Making(prediction) & and 
		
		(Category I) &  &  &  &  \\
		Partially Explicit Decision-Making(prediction) & & and 
		
		(Category II)& & & \\
		Explicit Decision-Making(prediction) &  &  & or
		
		(Category III)& & \\
		Explicit Decision-Making(prediction) &  &  &  & and 
		
		(Category IV) & \\
		Explicit Decision-Making(prediction) &  &  &  & & and 
		
		(Category V)\\
		\bottomrule 
	\end{tabular}
\end{table}

\section{Explainability-based Classification of Neuro-symbolic Methods}	
\subsection{Implicit Intermediate Representations and Implicit Decision Making(prediction)}	
This category includes 74 neuro-symbolic AI studies with three common characteristics, as Table 2 shows. First, they all use neural networks to extract features from data. However, the representations of these features cannot be directly processed by symbolic logic, so they need an intermediate representation to fill the gap. Secondly, the intermediate representation bridging the two is often latent vector embedding or combined with structured representation but is only partially explicit and directly human-readable. Third, most of the overall decision-making logic or prediction method is implicitly expressed through the weights and activation functions of the neural network. Some methods directly integrate symbolic logic into the decision-making process or provide an indirect understanding of decision-making logic by designing interpretable interfaces such as attention mechanisms and logic rule generators. However, decision-making logic still needs to be explained overall.

\begin{table}[ht]
	\centering 
	\caption{Implicit Intermediate Representations and Implicit Decision Making} 
	\label{tab:neuro_symbolic} 
    \begin{tabular}{|c|c|>{\centering\arraybackslash}p{6.7cm}|} 
		\hline 
		\multicolumn{2}{|c|}{Classifications} & Papers \\ 
		\hline
		\multirow{3}{*}{Fundamental Methods} & Sampling & \cite{ref8scassola2023conditioning}  \\ 
		\cline{2-3}
		& Graph Learning &  \cite{ref9werner2023knowledge,ref10ebrahimi2021neuro,ref11dold2022neuro,ref12chen2016multilingual,ref14singh2023neustip,ref15alshahrani2017neuro}      \\ 
		\cline{2-3}
		& Security &   \cite{ref16gopinath2018symbolic}     \\ 
		\hline
		\multirow{5}{*}{Logic and knowledge processing} & Logical Reasoning & \cite{ref17van2023nesi,ref18besold2017reasoning,ref19machot2023bridging,ref20dong2019neural,ref21zhu2023tgr,ref22zhu2023approximate,ref23odense2022semantic,ref24lazzari2024sandra,ref25li2024softened,ref26cingillioglu2022end,ref27fire2016learning,ref28ahmed2022semantic,ref29wang2019satnet,ref30tran2017unsupervised,ref31ying2023neuro}       \\ 
		\cline{2-3}
		& Probabilistic Reasoning & \cite{ref32gal2015bayesian,ref33franklin2020structured,ref34amado2023robust,ref35mnih2014neural,ref36marra2021neural,ref37riveret2020neuro,ref38marconato2024bears,ref39ahmed2023semantic,ref40stehr2022probabilistic,ref41furlong2023modelling}       \\ 
		\cline{2-3}
		& Neural Automata &  \cite{ref42uria2023invariants}     \\ 
		\cline{2-3}
		& Explainability &   \cite{ref43diaz2022explainable}    \\ 
		\cline{2-3}
		& Knowledge processing &   \cite{ref44cohen2020scalable,ref45lemos2020neural,ref46marconato2023neuro}     \\ 
		\hline
		\multirow{7}{*}{Application} & Question Answering &  \cite{ref47pinhanez2020using,ref48ma2019towards,ref49bosselut2021dynamic,ref50zheng2022jarvis,ref51barcelo2023neuro}      \\ 
		\cline{2-3}
		& Programming &   \cite{ref52devlin2017semantic,ref53davis2022neurolisp,ref54hasija2023neuro}     \\ 
		\cline{2-3}
		& Education &   \cite{ref55hooshyar2024temporal,ref56hooshyar2023augmenting}     \\ 
		\cline{2-3}
		& Fault diagnosis &  \cite{ref57saravanakumar2021hierarchical}      \\ 
		\cline{2-3}
		& Chemical &  \cite{ref58segler2018planning}      \\ 
		\cline{2-3}
		& Link prediction &   \cite{ref59rivas2022neuro}     \\ 
		\cline{2-3}
		& Emotion analysis &  \cite{ref60baran2022linguistic}      \\ 
		\hline
		\multicolumn{2}{|c|}{Mathematics/Symbolic Regression} &  \cite{ref61flach2023neural,ref62biggio2021neural,ref63podina2022pinn,ref64lample2019deep,ref65bendinelli2023controllable,ref66cai2017making,ref67long2019pde,ref68arabshahi2018combining,ref69mundhenk2021symbolic,ref70kubalik2023toward,ref71bahmani2024discovering}      \\ 
		\hline
		\multicolumn{2}{|c|}{Natural language processing} &  \cite{ref72karpas2022mrkl,ref73shakya2021student,ref74chrupala2019correlating,ref75hu2022empowering}      \\ 
		\hline
		\multicolumn{2}{|c|}{Computer vision} &  \cite{ref76cheng2023transition,ref77cingillioglu2021pix2rule,ref78mota2016shared}      \\ 
		\hline
		\multicolumn{2}{|c|}{Reinforcement learning} &  \cite{ref79dutta2023s,ref80yan2023point,ref81chitnis2022learning,ref82garg2020symbolic}      \\ 
		\hline
	\end{tabular}
\end{table}
\citet{ref45lemos2020neural} proposed a neural symbolic model for relational reasoning and link prediction on knowledge graphs. The model takes a subset of the knowledge graph as input and then uses two learned embedding layers to map entity types and relationships into real-valued vector space to initialize the representation of entities and relationships so that entity types and relationships passing through the embedding layer are converted into real-valued vectors that capture the underlying semantic features of entity types and relationships. Then, the graph neural network is used to update the vector representation of entities and relationships iteratively, and through the message-passing mechanism of GNN(Graph Neural Network), each entity, which is represented by a node and relationship, which is represented by an edge in the graph exchange information and update their adjacency relationships. Representation captures complex dependencies between entities and relationships, and each entity and relationship has a unique identity in the symbolic logic graph. Finally, a multilayer perceptron is used to decode the vector representation of the target relationship and convert it into the predicted relationship type. This representation  is a symbolic one that the neural network learns as a real-valued vector and converts to symbolic logic form to bridge the gap from data-driven feature extraction to symbolic logic reasoning. A core challenge faced by this research is the need for more consistency in the representation form. That is, while the neural network uses real-valued vectors to represent entities and relationships and captures a large amount of unstructured information in this form, the symbolic logic part relies on clearly defined symbols and rules to perform reasoning. This inconsistency necessitates the use of an intermediate representation to bridge the two. In this study, the author uses embedding vectors as an intermediate representation to bridge the deep learning feature expression and symbolic logic. However, these high-dimensional and dense vectors need to be more explicit and directly readable, which makes it impossible to observe and verify their correctness directly. At the same time, the decision-making logic of this method is also implicitly expressed through the weights and activation functions of the neural network, so although the model can perform complex inference tasks, it is difficult to understand and examine the specific reasons why the model makes a particular prediction.

\citet{ref39ahmed2023semantic} proposed a method of assuming constraint decomposition to solve probabilistic reasoning problems conditioned on the characteristics of network learning. This paper implements the process through a semantic reinforcement approach, i.e., building logic circuits, calculating the probabilities of logic constraints, and using the probabilities output by neural networks to estimate the likelihood of different logic states. Semantic reinforcement methods assume that individual sub-constraints in a logical constraint are independent of each other, so the probability of entire constraints can be approximated by multiplying the probabilities of its sub-constraints. Therefore, basic logic operation units such as AND and OR gates are implemented by constructing logical constraint calculation graphs. This construction process follows structural rules such as decomposability and determination. Therefore, these logic circuits can effectively calculate each probability of satisfying logical constraints with the given input. At the same time, based on probabilistic decomposition, this method quantifies how strong the dependence between two sub-constraints is under the given neural network characteristics by calculating Conditional Mutual Information. It uses this to evaluate which sub-constraints are independent assumptions that may lead to inaccuracies in probability calculations. The accuracy of overall probability calculation can be further improved by gradually relaxing the independence assumption between these sub-constraints and merging them into more complex constraints. Throughout the process, features in the form of real-valued vectors are extracted from the input data through neural networks. However, they do not directly correspond to the discrete symbolic representation used by symbolic logic. Therefore, to bridge this gap, neural networks need to convert the extracted features into an intermediate representation by knowledge compilation technology, which can compile logical formulas into calculable circuit structures. However, it is implicit and cannot be directly observed and verified. Second, the overall decision logic is implicitly expressed through the weights and activation functions of the neural network and is not entirely transparent to external observers. Although computing conditional mutual information and performing semantic reinforcement provides a way to iteratively improve the decision logic to optimize the processing of logical constraints, the explainability of overall decision process still needs improvement.

\citet{ref46marconato2023neuro} proposed a neural symbolic continuous learning method and an optimization strategy, COOL(concept-level continual learning), to solve the catastrophic forgetting problem in a series of neural symbolic tasks. First, the method extracts features from visual data. It then uses a separate neural network to extract high-level concepts from these features, such as color and shape in the CLEVR(Compositional Language and Elementary Visual Reasoning) dataset or number recognition in the addition task in the MNIST(Modified National Institute of Standards and Technology database) dataset, which are then generated in final predictions or decisions based on the constraints of given prior knowledge. In this process, symbolic rules exist as explicit mathematical and logical expressions and serve as prior knowledge K to define the relationship between concepts and the mapping rules between concepts and predicted results. When faced with new tasks, this method adapts to the new concept representation by updating the neural network, modifying or extending the logic rules to reflect the new knowledge, and achieving continuous learning based on the new concept representation. After the neural network extracts from the visual input, these features exist as vectors, and this method uses another neural network as an intermediate representation to convert these vectors into visual concepts, as logical symbols cannot process them further. However, the intermediate representation is implicitly expressed through weights and activation functions; they cannot be directly observed or tested. So, even though this method's reasoning and prediction process is explicitly based on mathematical formulas or logical expressions, it still needs to be fully interpretable.

\citet{ref41furlong2023modelling} proposed a method for realizing probability calculation through VSA (vector symbolic architecture) and SSP(spatial semantic pointers) applied to cognitive models and neural networks. This method usually takes observation data or features as input data, such as images, sounds, or other types of sensor data. These observations are first converted into points as SSPs in a high-dimensional vector space by converting data into continuous space, where each vector represents an observation. Then, use the bundling, binding, unbundling, and other operations defined in VSA to perform abstract symbolic logic processing on the encoded data, such as calculating unified distribution, conditional distribution, entropy, and mutual information, etc., and executing calculations of conditional probabilities through unbundling operations, or estimating entropy or mutual information through a specific neural network structure, which means that probability calculations can be performed directly on the SSP representation through the above symbolic operations, based on this probability information, decision logic or prediction can be constructed, such as selecting actions with the highest expected reward, updating belief states based on current observations, or other forms of high-level cognitive tasks. This method can support constructing models that simulate human or animal cognitive processes such as decision-making, learning, and memory in handling uncertainty and probabilistic reasoning. It is worth mentioning that this method uses a neural network to map input information into high-dimensional vectors and uses logical symbolic methods to process these features directly. Therefore, although these two expressions are the same, the method can only be classified as low explainable, as the logical reasoning operations are in the implicit space.

In this category, the form of intermediate representation and the inference or prediction process partly depend on the neural network's output. Therefore, even though part of the process may be combined with an explicit logical symbolic process, it still belongs to low explainability as all efforts have not outstripped the black box effect of neural networks.
\subsection{Partially Explicit Intermediate Representations and Partially Explicit Decision Making(prediction)}	
As Table 3 shows, this category involves 110 neuro-symbolic AI studies, and they have three common characteristics. First, they all use neural networks to extract features from data. However, the representations of these embeddings cannot be directly processed by symbolic logic, so they need an intermediate representation to fill the gap. Secondly, most intermediate representations are symbolic logic expressions, mathematical expressions, structured programs, logic circuits, probability distributions, virtual circuits, and virtual machine instructions. These representations are partially explicit and human-readable. Third, decision logic combines implicit representations from neural networks and explicit representations of symbolic logic, so it is partially explicit and readable.

\begin{table}[htbp]
	\centering
	\caption{Partially Explicit Intermediate Representations and Partially Explicit Decision Making}
	\label{tab:research_classification}
    \begin{tabular}{|c|c|>{\centering\arraybackslash}p{6.5cm}|}
		\hline
		\multicolumn{2}{|c|}{\textbf{Classification}} & \textbf{Papers} \\
		\hline
		\multicolumn{2}{|c|}{Representation learning/graph learning} & \cite{ref83sansone2023learning,ref84finzel2022generating,ref85niu2021perform,ref86raj2023neuro,ref87chaudhury2023learning,ref88zhu2022neural} \\ 
		\hline
		\multicolumn{2}{|c|}{Natural language processing} & \cite{ref89liang2016neural,ref90saha2021weakly,ref91zhang2023natural,ref92liu2023weakly,ref93chaudhury2021neuro,ref94pallagani2022plansformer,ref95chen2019neural,ref96arabshahi2021conversational,ref97chaudhury2023learning,ref98verga2020facts,ref99hwang2021comet,ref100verga2021adaptable,ref101hu2023chatdb,ref102liu2022neural,ref103zellers2021piglet,ref104alon2022neuro,ref105chen2020compositional,ref106schon2021negation,ref207ref3Galassi }\\
		\hline
		\multicolumn{2}{|c|}{Computer vision} & \cite{ref107feinman2020learning,ref108feinman2020generating,ref109khan2023neusyre,ref110sarkar2015early,ref111wu2022zeroc,ref112shindo2021neuro,ref113manigrasso2023fuzzy,ref114amizadeh2020neuro,ref115yu2022probabilistic,ref116asai2020learning,ref117su2022probabilistic,ref118alford2021neurosymbolic,ref119dang2020plans,ref120chen2023genome,ref121le2021scalable,ref122wang2023rapid,ref123agarwal2021end,ref124hsu2023ns3d,ref209apriceno2021neuro} \\
		\hline
		\multicolumn{2}{|c|}{Reinforcement learning} & \cite{ref125anderson2020neurosymbolic,ref126moon2021plugin,ref127nunez2023nesig,ref128garcez2018towards,ref129garnelo2016towards,ref130mitchener2022detect,ref131singireddy2023automaton,ref132landajuela2021discovering,ref133lyu2019sdrl,ref134hazra2023deep,ref135silver2022learning,ref136sun2021neuro,ref137yang2018peorl,ref138illanes2020symbolic,ref139balloch2023neuro,ref140akintunde2020verifying,ref141sharifi2023towards,ref142jiang2024multi,ref213illanes2020symbolic}\\
		\hline
		\multicolumn{2}{|c|}{Mathematics / symbolic regression} & \cite{ref143majumdar2023symbolic,ref144petersen2019deep,ref145d2022deep,ref146kim2020integration,ref147cranmer2020discovering,ref148qin2021neural,ref149gaur2023reasoning,ref150daniele2022deep,ref151tong2023neolaf}\\
		\hline
		\multirow{2}{*}{Logic and knowledge processing} & Concept/rule learning & \cite{ref152aspis2022embed2sym,ref153stammer2021right,ref154baugh2023neuro,ref208van2017linking}\\
		\cline{2-3} 
		& Logical reasoning & \cite{ref155cunnington2023ffnsl,ref156cunnington2022neuro,ref157tao2024deciphering,ref158tsamoura2021neural,ref159dragone2021neuro,ref160xie2022neuro,ref161zhang2018interpreting,ref162glanois2022neuro,ref163bonzon2017towards,ref164pan2023logic,ref165nye2021improving,ref166manhaeve2019deepproblog,ref167bennetot2019towards,ref168li2020closed}\\
		\hline
		\multirow{7}{*}{Applications} & Visual question answering & \cite{ref169mao2019neuro,ref170yi2018neural,ref171vedantam2019probabilistic,ref172siyaev2021neuro}\\
		\cline{2-3}
		& Medical & \cite{ref173jain2023reonto,ref174fadja2022neural,ref175han2021unifying}\\
		\cline{2-3}
		& Communication & \cite{ref176thomas2023neuro,ref177thomas2022neuro}\\
		\cline{2-3}
		& Programming & \cite{ref178hu2022fix,ref179tarau2021natlog,ref180daggitt2024vehicle,ref181arakelyan2022ns3}\\
		\cline{2-3}
		& Recommended system & \cite{ref182carraro2023overcoming,ref210lyu2022knowledge}\\
		\cline{2-3}
		& Security & \cite{ref183wang2018formal,ref212trapiello2023verification}\\
		\cline{2-3}
		& Others & \cite{ref184jia2023symbolic,ref185ashcraft2023neuro,ref186cosler2024neurosynt}\\
		\hline
		\end{tabular}
	\end{table}

\citet{ref144petersen2019deep} proposed a DSR(Deep Symbolic Regression) method to recover mathematical expressions from data. This method first represents each expression as a node sequence through symbolic expression trees, including mathematical operators such as addition, subtraction, multiplication, and division and operands such as x and constants. A RNN(Recurrent Neural Network) is then used to predict the next operator or operand based on the existing sequence. In this method, RNN predicts new nodes based on existing expression trees, and logical symbols exist as expression trees of mathematical expressions. The tree structure represents the logical relationship between operators and operands, such as which two numbers are added together or which number is divided by another number. Each time the RNN generates a complete expression, the expression is used to calculate the degree of fit on a specific data set, such as by calculating the difference between the expression and the actual data, and the fitting is feedback to the RNN to guide the adjustment of the subsequent expression generation process. In DSR, the weights, activation functions, and currently processed sequences inside the RNN act as an implicit intermediate representation, although this representation is not directly oriented to the end user, nor is it in any explicit or structured way. The symbolic form exists, but it connects the input sequence and the output prediction and is finally output explicitly in the form of a specific symbol in a mathematical expression. Hence, the process is partially explicit and interpretable. The logical decision-making part of DSR can also be divided into two steps: the explicit evaluation process and the implicit adjustment process. The former refers to completing the calculation of the fit or reward of the expression based on the comparison of the expression and the data. The latter is intended to provide feedback on the degree or reward to the RNN to guide the subsequent expression generation process, which relies on RNN gradient descent or other optimization algorithms and is achieved by adjusting the weight of RNN, so it is also a partially explicit process.

In Part I, we mentioned the NSPS(neuro-symbolic program search) method proposed by \cite{ref136sun2021neuro} to improve the automation level of autonomous driving system design by automatically searching and synthesizing neural symbolic programs. This method uses neural networks to extract rich information required for autonomous driving, such as vehicle speed, acceleration, and attitude, from the input data. However, it cannot be directly used for symbolic logic processing. Therefore, NSPS bridges the two by defining a DSL(domain-specific language), which contains basic primitives for driving attributes, such as vehicle speed and acceleration, and statements for implementing advanced priors. However, the method automatically selects and combines these operations through program search and is, therefore, partially explicit. However, the operations defined by DSL can directly reflect the logic of driving decisions, so they are also partially explicit. 

For the same reason, the final driving decision of the decision-making process is obtained by automatically searching and combining a given set of neural-symbolic operations. This process relies on the implicit expression of the neural network weights and activation functions. Although the symbolic operations provide a certain degree of interpretability, specific working methods of the underlying neural network model and program search algorithm are still implicit, which leads to the overall decision-making logic of NSPS being partially explicit.

\citet{ref84finzel2022generating} proposed using GNN(graph neural networks) to classify relational data and verify the GNN output by generating understandable explanations combined with ILP(inductive logic programming). This method first uses GNN to extract features from graph-structured data. These data include patterns in the Kandinsky pattern data set, which are composed of different geometric objects such as circles and triangles, as well as attributes such as shape, color, size, and location of the node. GNN aggregates the information of neighbor nodes through the connection relationships between them, updates the feature vectors of nodes, and identifies the graph structure that has the most significant impact on the classification results through interpreters such as GNN-Explainer. These graph structures and their importance scores are then converted into facts and rules in a Prolog program. In this case, symbolic logic exists in the form of ILP, which uses these transformed symbolic data as background knowledge to learn rules that describe the logic of classification decisions. These rules are not only based on structural features such as spatial relationships between nodes and node attributes such as color and shape but also comprehensively consider the importance scores of these features to generate understandable, logical rules that explain GNN classification decisions. The intermediate representation of this method converts the output of GNN into Prolog facts and rules suitable for ILP processing and builds a bridge from neural network features to symbolic logic. Therefore, this intermediate representation combines the two forms of latent vector embedding and Prolog rules and is partially explicit. Similarly, the decision-making logic in this method involves both two parts. The feature representation automatically learned by the neural network from the data is used as the input of the ILP, and it makes subsequent decisions based on this representation. Therefore, the two are partially explainable.

\subsection{Explicit Intermediate Representations or Explicit Decision Making(prediction)}	
There are 3 neuro-symbolic AI studies in this category, as Table 4 shows, and all of which have three characteristics. First, neural networks are used to extract features from data. However, the representations of these features cannot be directly processed by symbolic logic, so they must use intermediate representations to fill the gap. Second, either intermediate representations or overall decision logic is entirely explicit.

\begin{table}[ht]
	\centering 
	\caption{Explicit Intermediate Representations OR Explicit Decision Making}
	\label{tab:neuro_symbolic} 
	\begin{tabular}{
			>{\centering\arraybackslash}m{5cm} 
			>{\centering\arraybackslash}m{5cm}
		}
		\toprule 
		Classification & Papers \\
		\midrule 
		Entity linking & \cite{ref187jiang2021lnn}\\
		NVM based robotic manipulation & \cite{ref188katz2021tunable} \\
		Question and Answering & \cite{ref189kapanipathi2020leveraging} \\
		\bottomrule 
	\end{tabular}
\end{table}

\citet{ref187jiang2021lnn} proposed an entity link prediction method based on LNN, LNN-EL(Logical Neural Network-Entity linking) to solve the entity linking problem in short texts, detailed in the multimodal non-heterogeneous Neuro-Symbolic AI section in Part I of this article. This method converts a set of logical rules into the network structure of LNN in LNN. It accurately matches the mentioned entities by processing features extracted from text or knowledge graph data. As mentioned in Part I 3.4, LNN maps logical operations directly into neural networks. Its neural network and symbolic logic operate the same data type in the same representation space without an intermediate representation to bridge the two. At the same time, the activation state of each neuron or neuron group can directly correspond to the truth value state of a logical proposition, so the decision-making process of logical operations is more accessible to explain. However, for LNN-EL, using deep models in the feature extraction stage results in the intermediate representation between its output and the LNN input being only partially explicit. Therefore, even if the LNN network structure is used in the inference stage, it is only moderately interpretable.

\citet{ref189kapanipathi2020leveraging} proposed a NSQA(neural symbolic question answering) system based on semantic parsing and reasoning. The method first uses an AMR(Abstract Meaning Representation) to convert natural language questions into AMR graphs according to explicit linguistic rules. Then it uses a neural network model to identify and link entities and relationships, that is, entities mentioned in the text through entity linking match entities in the knowledge base, and use relationship links to match actions or relationships in the AMR graph with attributes or relationships in the knowledge base to convert the AMR graph into an intermediate representation corresponding to the entities and relationships. This representation is then further converted into logical queries, and a Logical Neural Network reasoner is used to infer based on the execution of these queries. The introduction of neural networks in the semantic parsing stage and entity and relationship linking stage of NSQA brings a certain degree of interpretability impact to the method. Although the output of these models can be directly verified by looking at the predictions on specific inputs, we still rate them as moderately explainable, considering that they are not as transparent as rule-based systems.

\citet{ref188katz2021tunable} proposed integrating high-level reasoning and low-level action control in robot manipulation tasks. This method is based on the neural virtual machine structure and can simulate the execution of a Turing-complete, purely symbolic virtual machine. The method first defines the NVM's (neural virtual machine) configuration, including layer size and connections between layers. It determines the source code of the symbolic algorithm to be executed on the neural virtual machine. Then, based on the initial configuration, an NVM instance with appropriate initial weights is automatically generated. A unique activity pattern vector represents each symbol in the source code. The compilation process of the source code is to apply a series of rules, such as fast storage-erasure learning, to the connection weight update of NVM, thereby encoding the symbolic algorithm into NVM. The compiled program can be executed by running the recursive dynamics of NVM. At each time step, the activity patterns of different layers in NVM represent the changes in the virtual machine state, and the weight matrix represents the program logic and data storage. Finally, the method uses a bidirectional lookup table called a "codec" to perform encoding and injecting symbolic input or extracting and decoding symbolic output at any time. This approach also means that the user can check the simulated machine state or provide input in symbolic form. It is worth mentioning that the decision-making logic of this method is implemented through changes in neural activation patterns, so the neural network's output uses the exact representation as the processing object of the logical symbol. However, although the two do not require an intermediate representation to bridge the two, their representation space is the high-dimensional vector space of the neural network. Hence, the overall decision logic needs to be more transparent and fully interpretable.
\subsection{Explicit Intermediate Representations and Explicit Decision Making(prediction)}	
There is 1 neuro-symbolic AI study in this category, which has three characteristics. The most significant difference between this classification and the previous one is that the intermediate representation and the overall decision logic are explicit. However, an intermediate representation is still needed to fill the gap between extracted features and symbolic processing.

\citet{ref190kimura2021neuro} et al. (2021) proposed a neural symbolic framework to solve reinforcement learning problems in text-based games. This method first extracts basic propositional logic from text observations obtained in the environment through a semantic parser, converting natural language text into a logical expression form. Then, external knowledge bases such as ConceptNet are used to understand the semantic categories of words in the text and refine the extracted propositional logic. Finally, the extracted propositional logic and the lexical category information obtained from ConceptNet are combined through the FOL(First Order Logic) converter and converted into first-order logical facts that reflect the specific characteristics and conditions of the game state. These logical facts are subsequently used as training input for LNN. LNN is trained to minimize the loss of logical contradictions. It learns symbolic rules from logical facts that can be mapped to action strategies, which are logical rules that can correspond to the best action strategies. Text-to-logic conversion is accomplished through a clear and explicit process, with the semantic parser explicitly mapping text information to a clear set of logical expressions. In some cases, the transformation process may involve enriching and supplementing these logical expressions with external knowledge bases, which generate more complex logical expressions based on associations between information in the text and data in the external knowledge base or consider previous observations or interaction history are used to construct the current logical state more accurately, so this process is also explicit. Likewise, these logical facts are logically reasoned and learned through LNN, operating based on clear, logical rules, so the logical decision-making process is explicit and explainable.
\subsection{Unified Representation and Explicit Decision Making(prediction)}	
There are 3 Neuro-Symbolic AI studies in this category, with three commonalities. First, although they use neural networks to obtain features, the neural network's output maintains the exact representation that can be processed by symbolic logic. Second, the overall decision logic is fully explicit and interpretable.

\citet{ref191riegel2020logical} proposed a LNN(logical neural network). As we mentioned in Section 3.4 of Part I, LNN directly interprets and operates logical operations by mapping each of its neurons to elements in the logical formula, which means that when the model processes information or makes decisions, its calculation process can be equivalent to performing a series of logical judgments, in which the output of each neuron not only represents the truth value of a logical proposition, and can also reflect how this truth value is derived from the input through logical operations, which is also the source of its high interpretability. In addition, each logical proposition in LNN is assigned a truth value range, and the uncertainty of the truth value of the proposition is captured and expressed through upper and lower bounds. For example, when faced with logical contradictions and incomplete knowledge, LNN will increase the uncertainty of the authenticity of certain propositions by expanding the truth value range of the proposition and finding a truth range adjustment solution that minimizes the contradiction. Similarly, when there is not enough information to determine the truth value of a proposition, LNN can reflect the uncertainty of the proposition by giving it a broader truth value range and dynamically adjust the true range as information is obtained to adapt to complex and dynamic changing information environment. Finally, unlike traditional single-task neural networks, LNN can simultaneously perform various logical reasoning tasks, such as theorem proving and fact derivation, demonstrating its high flexibility in logical processing and future development potential.

\citet{ref191sen2022neuro} proposed an LNN-based inductive logic programming method that learns interpretive rules from noisy, real-world data, which can generate interpretable logic rules as output based on structured input data. This method first uses a knowledge base containing facts and relations, as well as rules describing the form or structure of the target as input. Then, an LNN network is built based on the template to simulate logical connectives, where each node or neuron represents an expression or a logical rule combined with logical connectives. Because LNN allows the behavior of logical operations to be adjusted through the learning process, the facts in the knowledge base can be used as training data, and the parameters in the LNN can be correctly reflected through optimization algorithms such as back propagation and gradient descent, so that it can correctly reflect the logic relation between facts. Finally, the trained LNN can be converted into a set of logical rules that directly reflect the logical relationships in the input data. It can be used to reason, predict, or explain patterns in the data and is interpretable. This method applies LNN in the ILP direction, allowing it to handle the complexity and uncertainty in real-world data while improving learning efficiency and the quality of rules.

\citet{ref192sen2021combining} proposed using LNN to complete the knowledge base. This method first defines the goal of knowledge base completion based on known entities and relationships. In this process, two methods, CM(Chain of Mixtures) and MP(Mixture of Paths), are mainly considered. Then, Boolean logic is extended to the real-valued domain in a parameterized manner through LNN, LNN-$\wedge$is used to simulate the logical AND operation, NN-pred is used to represent the relationship mixture or path mixture in the rule body, and LNN is used to construct knowledge base completion rules based on the rule template, that is, for the CM method, each hop relationship is represented as a mixture of relationships. In contrast, a mixture of multi-hop relationship sequences is directly learned for the MP method. This method uses pre-trained knowledge graph embeddings to supplement rule learning for the shared path sparsity problem in knowledge-based completion tasks. In this way, the parameters in the LNN can be optimized by iterative training on a given knowledge base graph, minimizing the prediction error. Finally, human-interpretable rules are extracted from the LNN model and used to predict missing entities in the knowledge base.

In addition to the above studies, \citet{ref193he2024reduced,ref194arrotta2023neuro,ref195xu2018semantic,ref211arrotta2024semantic}proposed loss functions suitable for Neuro-Symbolic Learning, and \cite{ref196ahmed2022neuro} proposed a regularization method suitable for neuro-symbolic learning. However, the above studies did not improve the models’ interpretability.

\section{TRENDS}	
This study used the keywords ‘neuro-symbolic’, ‘neuro symbolic’, and ‘neuro symbolic learning’ to survey relevant research from 2014 to the present on Google Scholar and Research Gate. Although the search was conducted to the greatest extent, it is possible that some literature could not be included due to limitations. Therefore, the trend comparison below mainly aims to illustrate the evolution of research trends rather than provide an accurate total literature volume. According to the scope of this study, several apparent trends can be seen from 2014 to Feb of 2024: First, the number of papers has increased significantly, especially between 2020 and 2023; the number of papers published is on an upward trend, and its publication in 2023 has the most significant number of papers, 55 in total based on presenting data. These numbers reflect the primary growth in research interest and activity in neural symbolic systems.

\begin{figure}[h]
	\centering
	\includegraphics[width=\linewidth]{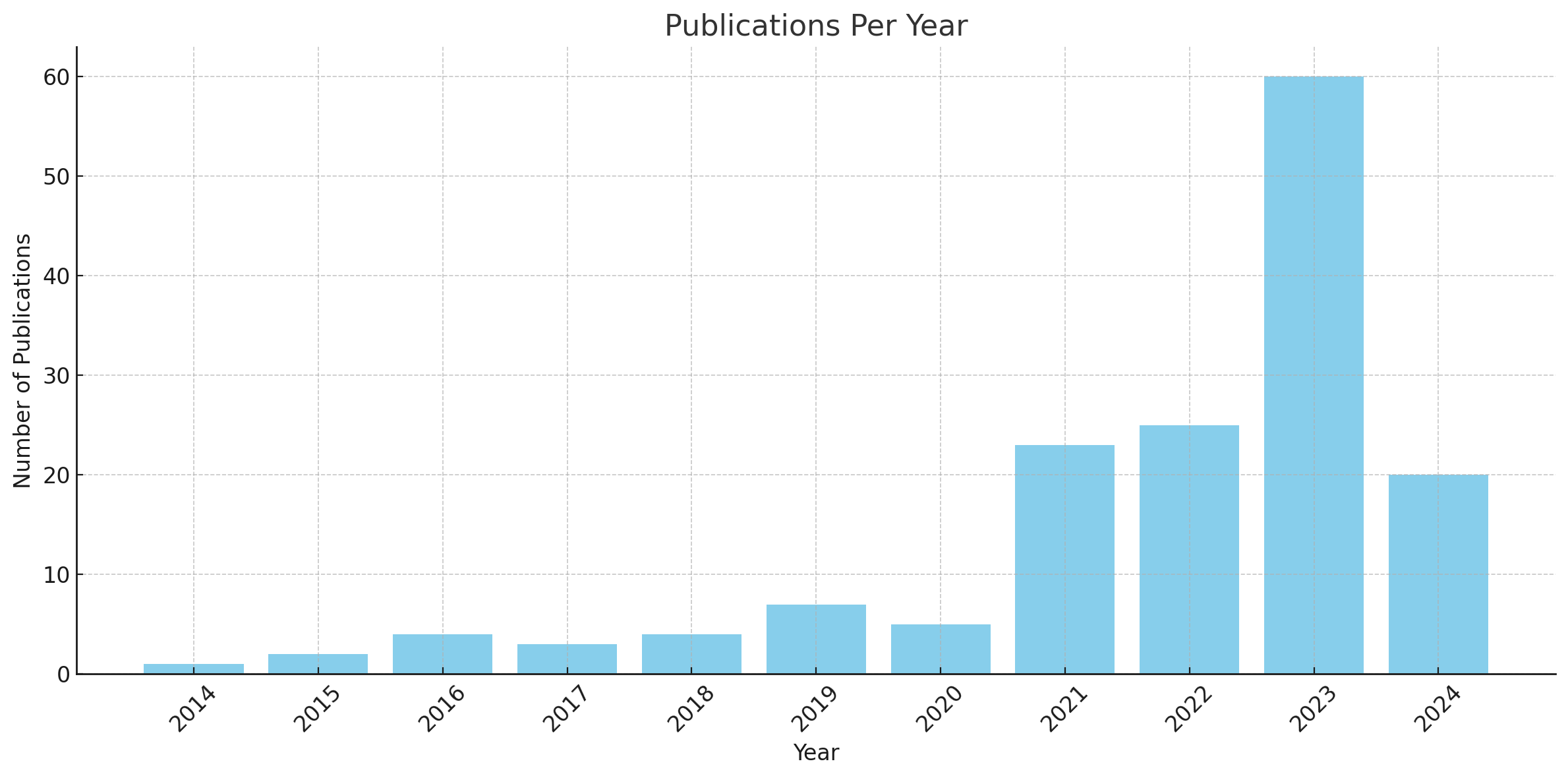}
	\caption{Annual publication trends}
\end{figure}
Secondly, among current neuro-symbolic methods, images and text are the most common input data types, reflecting their ubiquity and importance in studying neuro-symbolic systems.

There are relatively few explorations into numerical and mathematical expression processing, structured data processing, environment and state awareness, and multimodal data types. However, as Figure 1 shows, the number of neuro-symbolic studies on these four input types has continued growing since 2016.

In addition, the vast majority of research used unimodal and non-heterogeneous representation space, which shows that in neuro-symbolic system research, a single type of data, such as text, image, or structured data, is still more common in research objects. Only a few papers explore the representation space of unimodal heterogeneous, multimodal non-heterogeneous, and multimodal heterogeneous, showing that these directions are relatively new and may be future growth points for research.

Finally, the most significant papers had medium-low explainability and low explainability. This figure indicates that most research results in this area still need to be higher in explainability. Only a few research have achieved medium, medium-high, and high explainability, indicating that there are still significant challenges in improving the explainability of neural symbolic systems in the future.
\section{Major Research Challenges}	
\subsection{Unified Representations in Neural Networks and Symbolic Logic}	
The conversion of representation between neural networks and symbolic logic has constantly challenged neuro-symbolic learning. In the two most common situations of neuro-symbolic AI models, neural networks are utilized to augment the extraction of symbolic logical features by interpreting complex patterns or data structures, which enhances the effectiveness of traditional symbolic methods, or symbolic knowledge embedding is applied to provide rule constraints or reasoning logic for neural networks. However, these cooperations are usually optimized for specific tasks, which are difficult to adapt or transfer to new tasks or data sets. Therefore, retraining the neural network model or adjusting the symbolic logic rules are necessary when requirements change, which hints that specific combinations constrain the model's overall generalization ability. Alternatively, both diagrams can directly utilize the extracted features or learned knowledge when neural networks and symbolic logic modules use unified representation, bringing additional benefits, such as improving training and inference efficiency.

In addition, the two methods of cooperation between neural networks and symbolic logic mentioned above are synchronization processes of knowledge, except that they are inefficient and offline. However, applying the unified representation will make this synchronization more efficient. Specifically, this consistency avoids additional knowledge transformation steps and corresponding information loss, which can make the system more flexible and efficient, reducing reliance on large amounts of offline training data while decreasing the complexity of model update and maintenance in handling complex practical application problems such as multi-step reasoning tasks in natural language or image recognition.

Despite the benefits, designing an ideal unified representation still faces many obstacles. An ideal representation could capture the structural properties of symbolic logic while maintaining the essential patterns of the data. Such designs require a solid understanding of the data distribution and its latent relationship with the logical entities. For example, how to effectively associate the image features of an abstract definition with its symbolic definition "dangerous" in the same representation space means that we first need to have a deep insight into the intrinsic semantic 'identity' of the very definition between two different types of data, and then discover a reasonable spatial form to reflect this insight. In addition, knowledge alignment based on unified representation must explicitly verify new knowledge's reliability. These updates can not only effectively spread to the specific knowledge conversion process but also, to a certain extent, it is necessary to ensure the consistency of output before and after. However, no matter what, the process should be transparent and explainable. Nevertheless, it is more likely to be closer to solving the current conceptual stability problem of connectionism, as the conceptual structure based on unified representation is constrained by fixed logical rules in formation and update.

Finally, traditional symbolic logic reasoning relies on clearly defined logical rules and logical structures, even if they are represented in real-valued logic, while neural networks reason through fuzzy probability distributions. There are fundamental differences in reasoning mechanisms. Integrating these two reasoning architectures in a unified representation means exploring new reasoning frameworks and developing logical algorithms that simultaneously handle fuzzy and deterministic logic.
\subsection{Explainability and Transparency}	
Neural networks introduce inescapable black-box features and inferences in cooperation with symbolic learning. For the case where they are loosely coupled, the explainability of the neural network cannot get any improvement because even if logical symbols provide rules or constraints for the neural network as embedded vectors, the embedding process itself is not intuitive and requires the addition of complex logical symbol reasoning, and the increased complexity brought about by the interaction between the two. Alternatively, their semantic overlap can form at least partly complementary in explainability when neural networks and symbolic logic utilize the unified representation.
\subsection{Sufficient Cooperation}	
The current integration of neural networks and symbolic logic makes it difficult to avoid the intrinsic problems in both diagrams. For instance, the inexplicable inference, the training cost of neural networks, or the expression limitations and generalization problems of symbolic logic could be introduced to the integrated neural-symbolic model. System complexity and knowledge synchronization may all become new issues. 
One promising avenue for addressing these challenges is to develop a new model architecture. This architecture would apply an integration layer for the outputs of the neural network component and the symbolic logic component, potentially overcoming the limitations of the current integration. An elastic two-way learning mechanism could be utilized to synchronize their knowledge. However, it is crucial to consider explainability from the outset of the design process.
\section{Future Research Directions}	
\subsection{Unified Representations}	
In Part II of the review, we classify recent neuro-symbolic AI studies into five categories based on information diversity in processing and integrating, as well as the characteristics and capabilities of the representation space in expressing neural network and symbolic logic: unimodal non-heterogeneous, multi-modal non-heterogeneous, single-modal heterogeneous, multi-modal heterogeneous and dynamic adaptive neuro-symbolic AI. This classification method is a good starting point in understanding and characterizing neuro-symbolic AI's performance from the representation perspective. Therefore, unified representation may be one of the key directions in future breakthroughs, as it minimizes information loss and maximizes knowledge that represents efficiency for both diagrams.

In the meantime, representation space is another promising research direction. Most current neuro-symbolic methods use Euclidean space as the representation space. However, dealing with non-linear problems such as complex relationships, graph-structured data, and timing dependencies is difficult and may not be as efficient as non-Euclidean space. Euclidean space is just a case of non-Euclidean space. Although the latter still needs much exploration regarding specific design methods and combination with European space, it still provides valuable options for future neuro-symbolic AI research.
\subsection{Enhancing Model Explainability}	
The explainability of models has become an unavoidable challenge in artificial intelligence research. Although the neuro-symbolic AI method provides more substantial transparency than traditional AI to a certain extent, it still cannot meet the requirements for its application in critical fields. Explainability should be considered more during the design phase rather than an afterthought. However, considering the complexity of explainability, we must first establish the basis for explainability. For example, mathematics and physical laws can be regarded as correct standards to a certain extent, while human common-sense logic may be full of contradictions and logical fallacies. Similarly, explainability must be based on a relatively stable concept in neuro-symbolic AI to be more convincible. Therefore, verifying and updating knowledge in LLMs is also an open topic.

The explainability requirements for Neuro-Symbolic AI are essentially divided into two parts: process and result transparency. The former may be based on rigorous logic or formulaic arguments, which means that even if a neural network is used to generate symbols for logical reasoning, this process should be transparent and interpretable enough to verify correctness. The latter shows that some unique thinking habits should also be considered, such as common sense in providing contextual evidence for reasoning results.

Although it is unclear and has obstacles to understanding what is happening in the model, advances in computational neuroscience can still provide some inspiration for feature directions. For example, \citet{ref198castaneda2023probabilistic} found that human deductive reasoning and probabilistic reasoning processes rely on different neurocognitive mechanisms in the brain, and people can suppress prior knowledge for deductive reasoning according to task requirements. Not all reasoning processes can be attributed to probability. These insights may imply that our prior knowledge of deductive reasoning should be tailored to the specific situation. \citet{ref199trumpp2013masked} found that activity in sensory and motor areas during conceptual processing can also occur unconsciously. \citet{ref200popp2019brain} also found that words related to actions and sounds activate sensory and motor areas of the brain during conceptual processing. This result shows that the acquisition of conceptual knowledge relies on reproducing sensory and motor brain networks, which supports the view of ground cognitive theory. Designing neuro-symbolic AI with a stable working or knowledge memory structure may inspire brain function. \citet{ref201belekou2022paradoxical} found significant differences in brain activation patterns when processing paradox reasoning and deductive reasoning tasks. \citet{ref202coetzee2022dissociating} found that logical reasoning in the adult brain may be separated from language processing. This research may inspire us to consider more flexible reasoning paths in neuro-symbolic AI or even dynamically configurable task reasoning methods.
\subsection{Ethical Considerations and Social Impact}	
Ethical and social considerations are another inescapable problem. If a high portion of our content will come from generative AI in the future, then the significance of this content will be far beyond the scope of being measured by credibility alone. All moral requirements of today's society, such as fairness and justice, privacy protection, prejudice and discrimination, environmental ethics, technological ethics, humanitarianism, or even religion, should be included in the evaluation criteria of AI algorithms. Just as the research by \cite{ref203van2023cleaning} shows, the significant correlation between computational models and the human brain does not prove that they have been implemented. As with the same process, indirectly interpreting the output of black-box AI does not meet process transparency requirements, so we may need designs that can fundamentally explain the generation process. Some research points out that neural-symbolic AI can provide more significant help in this regard by providing more rules and constraints for processing neural networks, such as \cite{ref204bello2023computational} proposed the application of the BDI(belief-desire-intention) model and reinforcement learning method to simulate how subjects internalize norms and make decisions consistent with human moral judgment, providing effective modeling and examples in Neuro-Symbolic AI in the future. It provides a possible inspiration for complex moral behavior.

In summary, we discussed the possible future development directions of neuro-symbolic AI from three aspects: unified representation, enhancing model explainability, ethical considerations, and social impact. These three aspects represent expectations for Neuro-Symbolic AI at three different levels. The representation space determines the technical foundation and the extent to which we can achieve reasonable explainability and transparency requirements, affecting the overall application prospects of neuro-symbolic AI and its impact on ethics and society.

\section{Acknowledgment}
I would like to thank my supervisor VS Sheng,whose expertise was invaluable in formulating the research questions and methodology. Your insightful feedback pushed me to sharpen my thinking and brought my work to a higher level.
\bibliographystyle{ACM-Reference-Format}
\bibliography{reference3}

\end{document}